\documentclass[journal,article,pdftex,moreauthors]{Definitions/mdpi}
\usepackage{etoolbox}
\makeatletter
\AtBeginDocument{%
  \@ifpackageloaded{lineno}{%
    \nolinenumbers            % apaga si ya está encendido
    \renewcommand{\linenumbers}{} % neutraliza llamadas posteriores
  }{}
}
\makeatother

\firstpage{1} 
\pubvolume{1}
\issuenum{1}
\articlenumber{0}
\pubyear{2025}
\copyrightyear{2025}
\datereceived{ } 
\daterevised{ } 
\dateaccepted{ } 
\datepublished{ } 
\hreflink{https://doi.org/} 

\Title{Low-Cost Open-Source Ambidextrous Robotic Hand with 23 Direct-Drive servos for American Sign Language Alphabet }
\TitleCitation{VulcanV3: Open-Source Ambidextrous Direct-Drive Hand for ASL}

\Author{Kelvin Gonzalez Amador $^{1}$\orcidA{}}
\AuthorNames{Kelvin Gonzalez Amador}

\isAPAStyle{%
       \AuthorCitation{Gonzalez Amador, K.}
         }{%
        \isChicagoStyle{%
        \AuthorCitation{Gonzalez Amador, Kelvin}
        }{
        \AuthorCitation{Gonzalez Amador, K.}
        }
}

\address{%
$^{1}$ \quad Independent Researcher, Los Angeles, CA, USA;\\
}

\corres{Correspondence: krobotics26@gmail.com;}

\abstract{Accessible communication through sign language is vital for deaf communities, yet robotic solutions are often costly and limited. This study presents VulcanV3, a low-cost, open-source, 3D-printed ambidextrous robotic hand capable of reproducing the full American Sign Language (ASL) alphabet (52 signs for right- and left-hand configurations). The system employs 23 direct-drive servo actuators for precise finger and wrist movements, controlled by an Arduino Mega with dual PCA9685 modules. Unlike most humanoid upper-limb systems, which rarely employ direct-drive actuation, VulcanV3 achieves complete ASL coverage with a reversible design. All CAD files and code are released under permissive open-source licenses to enable replication. Empirical tests confirmed accurate reproduction of all 52 ASL handshapes, while a participant study (n = 33) achieved 96.97\% recognition accuracy, improving to 98.78\% after video demonstration. VulcanV3 advances assistive robotics by combining affordability, full ASL coverage, and ambidexterity in an openly shared platform, contributing to accessible communication technologies and inclusive innovation.}


\begin{document}

\begin{figure}[H]
    \centering
    \includegraphics[width=0.55\linewidth]{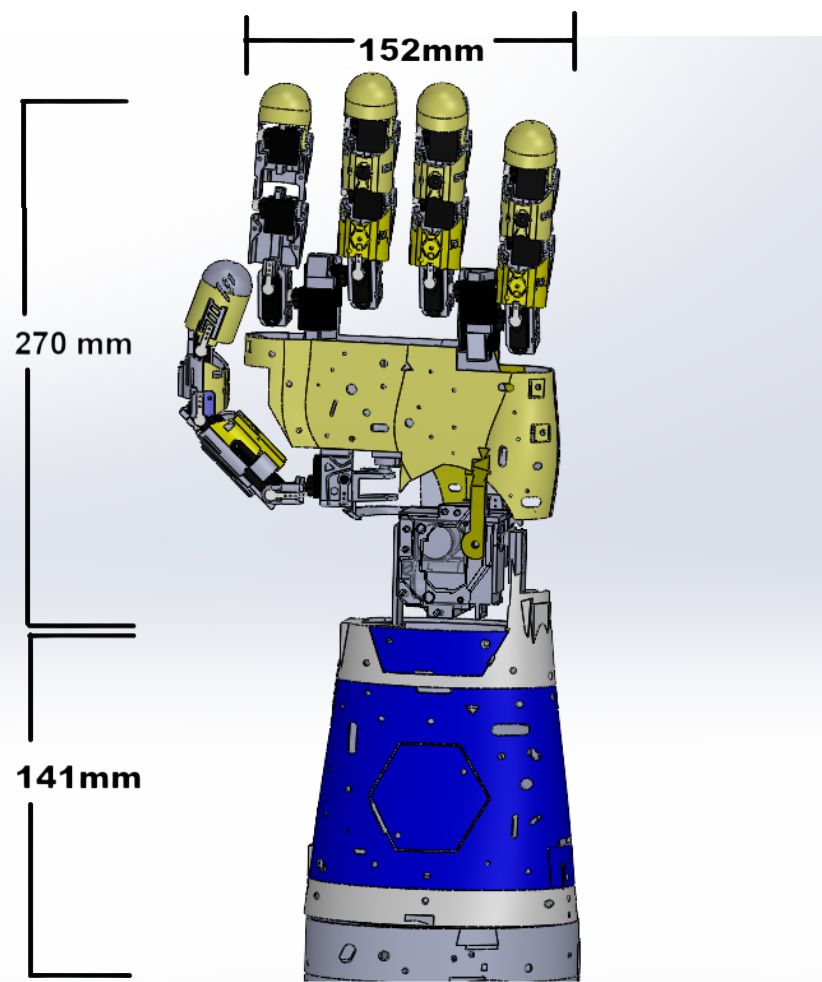}
    \caption{SolidWorks 3D CAD design of VulcanV3.}
    \label{fig:cad}
\end{figure}

%=================================================================
\section{Introduction}
Effective communication is fundamental to human interaction, yet for millions of deaf individuals, barriers to accessible sign-language solutions persist \citep{ref-Kushalnagar-2017}. American Sign Language (ASL) relies on precise handshapes and wrist orientations to convey meaning \citep{ref-Emmorey-2023}. Research prototypes have explored automated sign-language systems, from tactile-ASL devices such as TATUM \citep{ref-Johnson-EMBC-2021,ref-Johnson-Thesis-2021} to low-cost, locally-sourced servo--tendon hands \citep{ref-Adeyanju-2023}, humanoid platforms that generate sentences in sign \citep{ref-Gago-2019}, and fingerspelling robots like Project Aslan \citep{ref-Aslan-2017}, SignBot \citep{ref-SignBot-2019}, RoboTalk \citep{ref-RoboTalk-2019}, and single-hand ASL demonstrators \citep{ref-Gul-2021,ref-VulcanV2-Video}. However, most rely on tendon transmissions and are unimanual, limited to single-hand operation, unlike ambidextrous systems; few explore fully direct-drive hands or ambidextrous execution, and those that do (e.g., commercial, and prototypes dexterous hands) are not targeting ASL coverage \citep{ref-Tesollo-2025, ref-Akyurek-2014}.

A recent overview of main drive methods used in humanoid robotic upper limbs indicates that direct-drive architectures remain the least used both in research and industry compared to tendon/cable, geared, or pneumatic approaches \citep{ref-DriveMethods,ref-Akyurek-2014}. Unlike most humanoid upper-limb systems, which rely on tendon-based or geared mechanisms, direct-drive architectures offer reduced mechanical complexity and improved torque accuracy \citep{ref-Taylor-1989}. Addressing this gap, this work introduces VulcanV3, a low-cost, fully open-source humanoid hand with 23 in-hand direct-drive servos, ambidextrous operation, and complete ASL coverage in both configurations (52 signs). Compared to prior art (e.g., TATUM’s single-hand alphabet \citep{ref-Johnson-EMBC-2021}, low-cost tendon designs \citep{ref-Adeyanju-2023}, or partial-letter systems \citep{ref-Piascik-2025,ref-SignBot-2019}) VulcanV3 substantially extends capability by combining full ASL with ambidexterity and fully direct-driven fingers and palm/wrist in a reproducible, openly shared platform.

%=================================================================
\section{Materials and Methods}
All 3D CAD, servo mappings, and Arduino code are open-source (CAD: CC-BY-SA; code: MIT) and are available in the publicly archived dataset on the Hackaday project page (see Data Availability section).

\begin{figure}[H]
    \centering
    \includegraphics[width=0.5\linewidth]{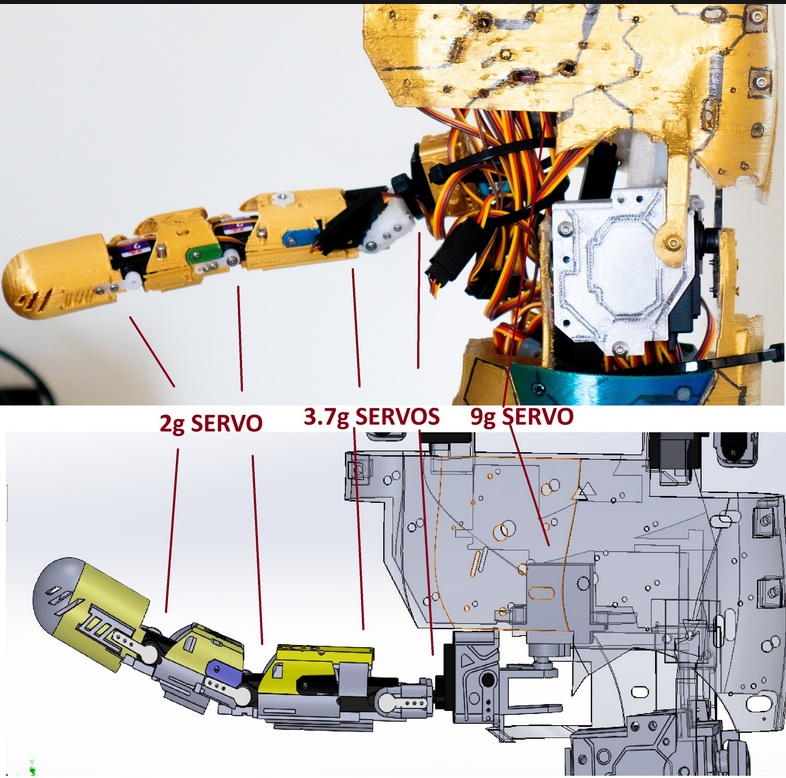}
    \caption{Thumb with 5 DOF.}
    \label{fig:thumb}
\end{figure}

\subsection{Mechanical Design}
The hand was designed in SolidWorks (Fig. 1) to achieve a humanoid structure with a wide range of motion for fingers, wrist, and forearm (Table~\ref{tab:rom}). The design emphasizes direct-drive and ambidextrous functionality. 

\begin{table}[H]
    \centering
    \caption{Degrees of amplitude for each part and movement type.}
    \label{tab:rom}
    \begin{tabular}{lcc}
        \toprule
        \textbf{Component} & \textbf{Movement} & \textbf{Range of Motion (°)} \\
        \midrule
        Forearm & Pronation--Supination & 270 \\
        Wrist & Radial--Ulnar Deviation & 145 \\
        Wrist & Flexion--Extension & 190 \\
        All Finger Phalanges & Flexion--Extension & 180 \\
        Index and Pinky & Abduction--Adduction & 100 \\
        Middle and Ring & Abduction--Adduction & 45 \\
        Thumb base & Abduction--Adduction & 195 \\
        Thumb & Pronation--Supination & 180 \\
        \bottomrule
    \end{tabular}
\end{table}

\subsection{Actuation}
VulcanV3 uses 23 direct-drive servo actuators within the hand: 21 micro servos (four per finger---two 2\,g and two 3.7\,g microservos---with the thumb using five including one 9\,g unit (Fig. 2)), plus 2 MG996R servos for palm/wrist. An optional 45kg forearm servo (270°) is not included in the 23 count. (Table~\ref{tab:rom2})

\begin{table}[H]
    \centering
    \caption{Specifications of servo motors used for the robotic hand, including dimensions, torque, weight, and operating speed}
    \label{tab:rom2}
    \begin{tabular}{lcccc}
        \textbf{Model} & \textbf{Dimensions (mm)} & \textbf{Torque} & \textbf{Weight (g)} & \textbf{Speed (sec/60°)} \\
        \hline
        C02CLS & 16.0 × 8.2 × 14.5 & 110 @ 4.2V & 2.2 & 0.060 @ 4.2V \\
        C037CLS & 20 × 8.5 × 17 & 550 @ 5.0V & 3.8 & 0.060 @ 5.0V \\
        MG996R & 40.7 × 19.7 × 42.9 & 11.0 kg-cm @ 6.0V & 55 & 0.17 @ 6.0V \\
        MG90S & 22.8 × 12.2 × 28.5 & 2.0 kg-cm @ 4.8V & 13.6 & 0.10 @ 4.8V \\
        45KG & 40 × 20 × 54 & 51 kg-cm @ 8.4V & 70 & 0.10 @ 8.4V \\
        \hline
    \end{tabular}
\end{table}

\subsection{Materials}
All fingers were printed with silk PLA on a standard FDM 3D printer; palm and forearm were printed in Nylon to improve structural rigidity.

\subsection{Electronics and Control}
The control hardware includes an Arduino Mega, two PCA9685 16-channel PWM modules, and a 5\,V supply. (Fig. 3). 

\begin{figure}[H]
    \centering
    \includegraphics[width=0.5\linewidth]{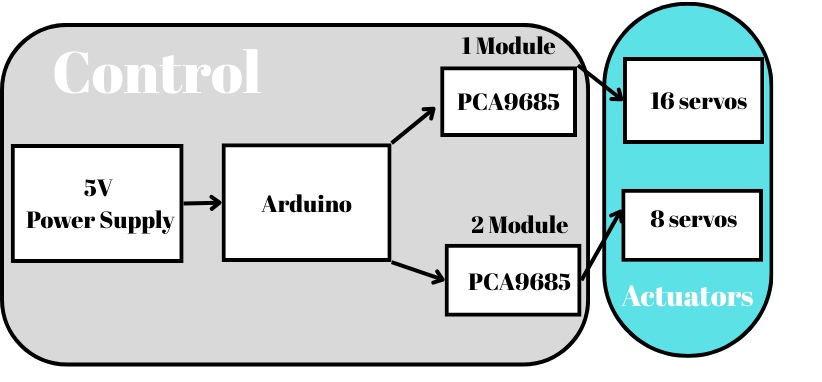}
    \caption{Electric control system.}
    \label{fig:control}
\end{figure}

\subsection{Programming and Mapping}
VulcanV3 was programmed using the Arduino IDE, with each ASL letter mapped to specific servo angles for both right- and left-hand configurations \citep{ref-Handspeak-2025} (Fig. 4). The system executes a complete sequence, starting from an initial position, performing the entire ASL alphabet in the right-hand configuration, returning to the initial position, and then reproducing the alphabet in the left-hand configuration (Fig. 5). 

\begin{figure}[H]
    \centering
    \includegraphics[width=1\linewidth]{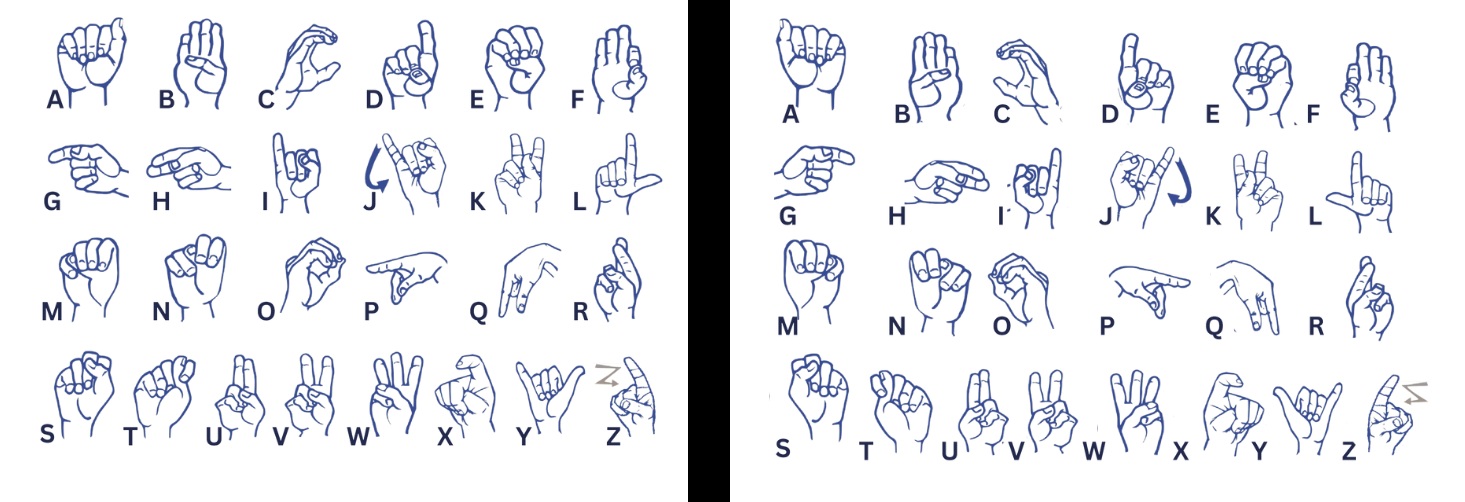}
    \caption{ASL alphabet Handshapes for both hand configurations.}
    \label{fig:asl}
\end{figure}

\begin{figure}[H]
    \centering
    \includegraphics[width=0.75\linewidth]{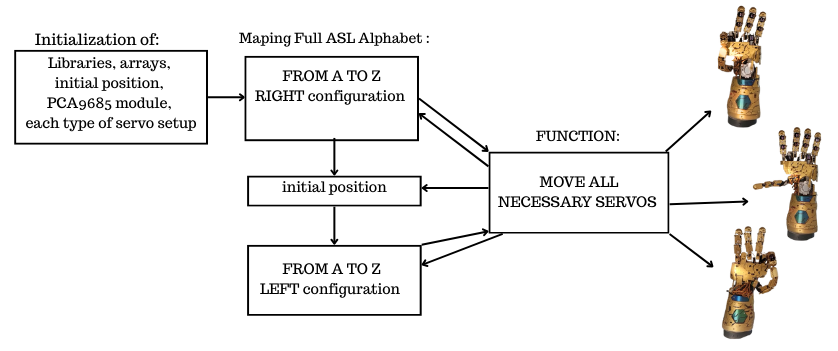}
    \caption{Programming flow chart.}
    \label{fig:flow}
\end{figure}

%=================================================================
\section{Results}
We evaluated VulcanV3's performance through empirical testing and a quantitative user validation study, following methodologies comparable to TATUM’s visual recognition validation \citep{ref-Johnson-EMBC-2021,ref-Johnson-Thesis-2021} and low-cost tendon-based prototypes \citep{ref-Adeyanju-2023}. A randomized handshape generator was implemented in the Arduino IDE to select and execute ASL letters in random order, enabling robust evaluation of recognition accuracy across both right- and left-hand configurations without prior participant exposure \citep{ref-Handspeak-2025}. The empirical test assessed the system’s ability to form intended ASL handshapes, while the user study measured recognition accuracy across varying levels of ASL expertise.

\subsection{Empirical Testing}
The system cycled through all 52 signs (26 letters $\times$ 2 configurations, right- and left-hand). Visual inspection against standard ASL diagrams \citep{ref-Handspeak-2025} confirmed a 100\% formation success rate for all letters (Fig. 6) and (Fig. 7), demonstrating precise servo angle mappings. Unlike prior unimanual or tendon-based systems \citep{ref-Adeyanju-2023,ref-SignBot-2019}, VulcanV3’s direct-drive architecture with 23 servos enabled accurate reproduction of complex handshapes, including ``R'' (requiring crossed fingers), ``G'', ``H'', ``P'', and ``Q'' (requiring wrist and forearm movements), and ``J'' and ``Z'' (requiring specific spatial trajectories). A demonstration video illustrating VulcanV3’s performance is accessible through the Hackaday project page (see Data Availability section).

\begin{figure}[H]
    \centering
    \includegraphics[width=1\linewidth]{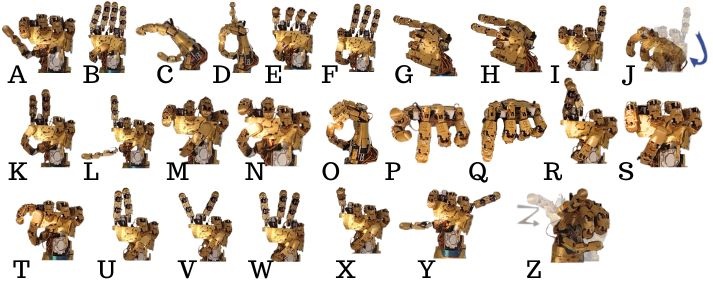}
    \caption{Left-hand configuration results.}
    \label{fig:left}
\end{figure}

\begin{figure}[H]
    \centering
    \includegraphics[width=1\linewidth]{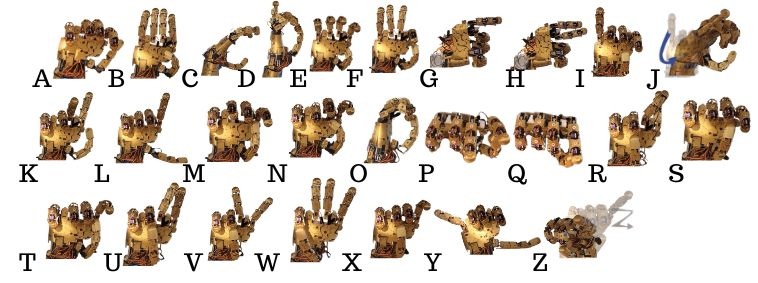}
    \caption{Right-hand configuration results.}
    \label{fig:right}
\end{figure}

\subsection{Quantitative User Validation}
Thirty-three participants, grouped by experience in ASL (19 with $>$ 10 years, 6 with $<$ 10 years, 3 teachers in ASL, and 5 with little knowledge in ASL), identified letters and configurations of signs presented in random order. The randomized handshape generator ensured unbiased testing by varying the sequence of the 52 signs. Table~\ref{tab:accuracy} summarizes the recognition accuracy per letter for the right- and left-hand configurations, achieving an overall recognition across all 52 signs and 33 participants was 96.97\% (1664/1716 correct). The accuracy was highest among teachers and users $>$ 10 years of age (near 100\%); most confusions (for example, M / N) occurred among users with little experience. After viewing a short demonstration video, a second test reached 98.78\% (1695/1716 correct).

\begin{table}[H]
    \centering
    \caption{Recognition accuracy for each ASL letter in the first test (no prior exposure), by configuration (\%).}
    \label{tab:accuracy}
    \begin{tabular}{lcc@{\hskip 1cm}lcc}
        \toprule
        \textbf{Letter} & \textbf{Left} & \textbf{Right} & \textbf{Letter} & \textbf{Left} & \textbf{Right} \\
        \midrule
        A & 93.94 & 100.00 & N & 81.82 & 75.76 \\
        B & 100.00 & 100.00 & O & 100.00 & 100.00 \\
        C & 100.00 & 100.00 & P & 90.91 & 96.97 \\
        D & 96.97 & 96.97 & Q & 93.94 & 93.94 \\
        E & 100.00 & 100.00 & R & 96.97 & 100.00 \\
        F & 100.00 & 100.00 & S & 81.82 & 93.94 \\
        G & 100.00 & 100.00 & T & 93.94 & 96.97 \\
        H & 96.97 & 100.00 & U & 100.00 & 100.00 \\
        I & 96.97 & 100.00 & V & 100.00 & 100.00 \\
        J & 100.00 & 100.00 & W & 100.00 & 100.00 \\
        K & 100.00 & 100.00 & X & 96.97 & 100.00 \\
        L & 100.00 & 100.00 & Y & 100.00 & 100.00 \\
        M & 87.88 & 84.85 & Z & 100.00 & 100.00 \\
        \bottomrule
    \end{tabular}
\end{table}

%=================================================================
\section{Discussion}
The results compare favorably with prior systems (Fig. 8): TATUM’s visual recognition rate ($\sim$94.7\%) with a single-hand alphabet \citep{ref-Johnson-EMBC-2021,ref-Johnson-Thesis-2021}, low-cost tendon-based hands reporting $\sim$78.4\% for letters \citep{ref-Adeyanju-2023}, partial-letter demonstrators \citep{ref-Piascik-2025,ref-SignBot-2019}, and fingerspelling arms like Aslan \citep{ref-Aslan-2017}. In contrast, VulcanV3 combines full ASL coverage with ambidexterity and fully direct-driven fingers/palm--wrist. This is noteworthy given that direct drive remains the least-used drive method in humanoid upper-limb actuation \citep{ref-DriveMethods}, both in research and industry.

\begin{figure}[H]
    \centering
    \includegraphics[width=1\linewidth]{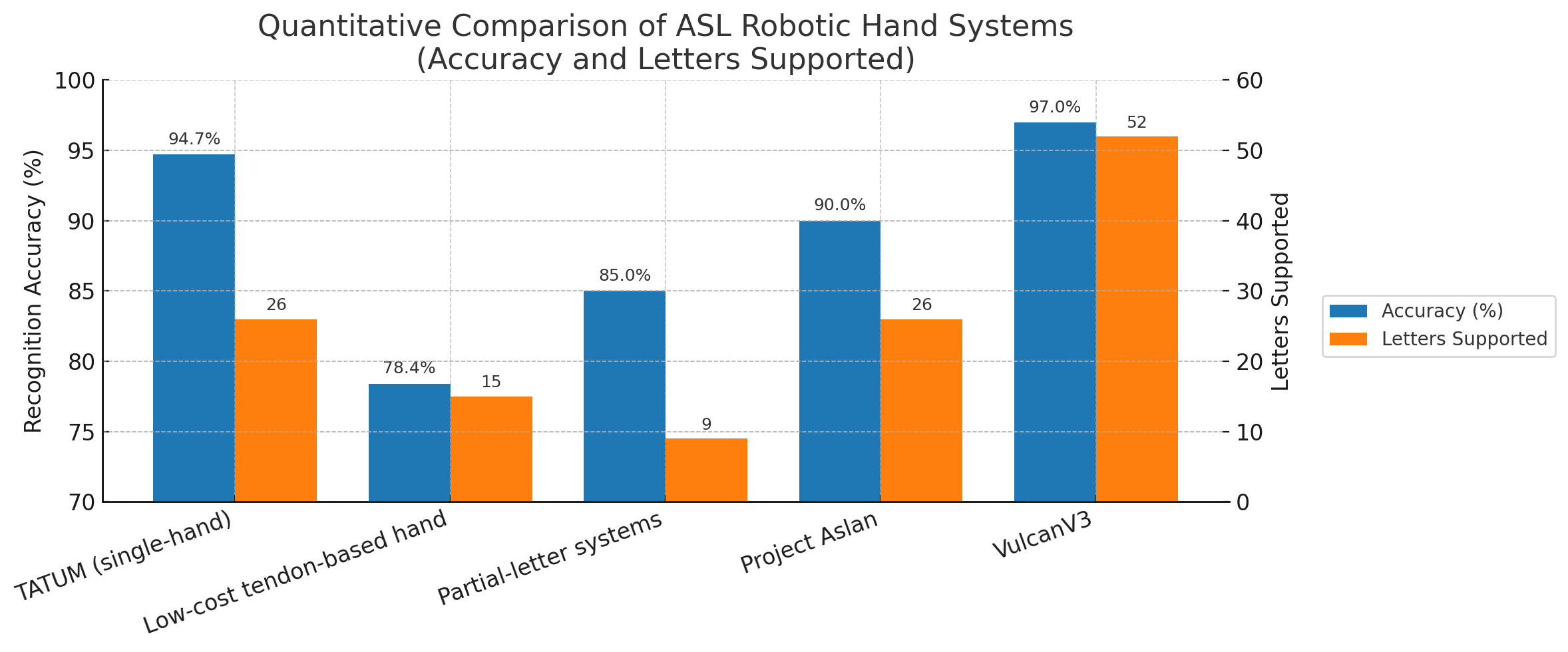}
    \caption{Quantitative comparison of ASL robotic hand systems. 
    The chart contrasts recognition accuracy (\%) and the number of letters supported by different approaches. 
    VulcanV3 achieves both the highest recognition accuracy (96.97\%, improving to 98.78\% after video demonstration) and complete coverage of the ASL alphabet in both left- and right-hand configurations (52 letters).}
    \label{Graphic comparison}
\end{figure}

Qualitative feedback from the 33 participants highlighted VulcanV3’s anthropomorphic design, and no participants reported safety concerns regarding its movements. However, some users with limited ASL experience struggled to differentiate visually similar handshapes, such as M'' and N'', P'' and Q'', and S'' and T'', which differ primarily in thumb placement or finger orientation. These confusions could be mitigated mechanically by extending the thumb length or adding an additional degree of freedom to enhance differentiation, as well as visually by painting the fingers a different color from the palm to improve contrast.

Despite these promising results, several limitations should be acknowledged. First, the system relies on low-cost servomotors, which may be prone to mechanical wear, audible noise, and reduced durability over prolonged use. These components also impose restrictions on movement speed and precision, requiring higher maintenance compared to industrial-grade actuators. Such factors may limit long-term deployment in demanding real-world environments. Furthermore, the hand currently focuses on alphabetic fingerspelling rather than continuous signing, and integration with multimodal systems such as vision-based recognition or speech-to-sign translation remains an open area for future work.

From an application perspective, VulcanV3 demonstrates potential beyond experimental validation. Its affordability and open-source nature make it a practical tool for educational institutions, research labs, and community programs seeking to increase awareness of sign language and assistive robotics. The design could serve as a teaching platform for engineering students, a demonstrator for public science outreach, or a foundation for low-cost assistive devices in contexts where commercial robotic hands are prohibitively expensive. By lowering barriers to replication and adaptation, VulcanV3 provides a pathway for advancing inclusion, accessibility, and innovation in both academic and societal domains.

%=================================================================
\section{Conclusions and future work}
This work introduces VulcanV3, an affordable, open-source, and ambidextrous robotic hand designed to facilitate American Sign Language (ASL) fingerspelling for deaf communities. Built from 3D-printed components and powered by 23 direct-drive servos in the hand plus one additional forearm servo, VulcanV3 demonstrates precise reproduction of all 52 ASL signs (26 letters in both right- and left-hand configurations). Empirical testing confirmed 100\% successful formation of all handshapes, while a user validation study with 33 participants achieved an overall recognition accuracy of 96.97\%, increasing to 98.78\% after exposure to a demonstration video. These findings highlight VulcanV3’s potential as a replicable, low-cost platform for education, research, and community outreach, aligning with global accessibility goals such as the UN Sustainable Development Goal 9 (Industry, Innovation, and Infrastructure). Future work will focus on visual enhancements (color coding), refined joint angles, and optional degrees of freedom to improve letter differentiation, as well as integration into a full humanoid arm capable of continuous sign communication while preserving low cost, simplicity, and ambidextrous operation.

%=================================================================

\begin{table}[H]\centering
\caption{Comparison of referenced robotic hands/arms for sign language.}
\begin{tabular}{p{5.0cm}p{1.6cm}p{1.1cm}p{1.2cm}p{2.2cm}}
\toprule
\textbf{System (Ref.)} & \textbf{Actuators} & \textbf{Direct Drive} & \textbf{Ambidex.} & \textbf{Coverage} \\
\midrule
TATUM (EMBC’21) \cite{ref-Johnson-EMBC-2021} & 15 & No & No & ASL 26 \\
TATUM (Thesis’21) \cite{ref-Johnson-Thesis-2021} & 17 & No & No & ASL 26 \\
Locally-sourced Hand \cite{ref-Adeyanju-2023} & 8 & No & No & 15 + 1--9 + 16 words \\
AI Arm (Teaching ASL) \cite{ref-Piascik-2025} & 13 & No & No & 9 letters \\
Project Aslan \cite{ref-Aslan-2017} & 16 & No & No & Alphabet + numbers \\
TEO Humanoid (UC3M) \cite{ref-Gago-2019} & n/s & No & No & LSE sentences \\
SignBot (MSL) \cite{ref-SignBot-2019} & 12 & No & No & 7 + 0--10 + phrases \\
RoboTalk \cite{ref-RoboTalk-2019} & 6 & No & No & DGS alphabet + 1--20 + \textasciitilde50 words \\
ASL Hand (Turkey) \cite{ref-Gul-2021} & 7 & No & No & 26 letters + 9 numbers\\
TESOLLO DG-5F \cite{ref-Tesollo-2025} & 20 & Yes & No & N/A \\
Vulcan V2 (video) \cite{ref-VulcanV2-Video} & 17 & Yes & No & 8 letters + 5 numbers \\
Ambidextrous Pneumatic Hand \cite{ref-Akyurek-2014} & 18 & No & Yes & N/A 
\\
\textbf{VulcanV3 (this work)} & \textbf{24 -- 23 in the hand} & \textbf{Yes} & \textbf{Yes} & \textbf{ASL 26 x2 (52) }\\
\bottomrule
\end{tabular}
\end{table}

%=================================================================

\dataavailability{All data supporting the reported results, including empirical testing and quantitative user validation results for the VulcanV3 robotic hand, are available at the publicly archived dataset on the Hackaday project page: \url{https://hackaday.io/project/203847-ambidextrous-23-direct-drive-humanoid-robotic-hand}.}

\section*{Use of Artificial Intelligence Tools}

Artificial intelligence tools (ChatGPT, OpenAI, 2025) were employed in the preparation of this manuscript. 
Specifically, AI assistance was used to (i) support the search and refinement of bibliographic references, 
(ii) correct minor grammatical errors and improve clarity of expression, and (iii) generate comparative chart. All content was carefully reviewed, validated, and approved by the author 
to ensure accuracy and scientific integrity.

\funding{This research received no external funding.}

\informedconsent{This study involved 33 human participants who voluntarily assessed the robotic hand's ASL letter recognition accuracy. The evaluation consisted of non-invasive visual observation and verbal feedback, posing no physical or psychological risk. As such, the study was deemed minimal risk, and no formal ethical approval was required under local regulations applicable to independent research [e.g., "in accordance with the ethical principles for human research outlined by the Declaration of Helsinki" or "under U.S. federal guidelines for minimal-risk studies (45 CFR 46)"]. All participants provided informed consent prior to participation, and their anonymity was maintained throughout the study. Data collection complied with general research ethics principles, ensuring voluntary participation and the right to withdraw at any time.}

\acknowledgments{The author have reviewed and edited the output and take full responsibility for the content of this publication.}

\reftitle{References}

\end{document}